\documentclass[wcp]{jmlr}
\usepackage{dsfont}

\title{Automatic clustering of a network protocol with
  weakly-supervised clustering}
\author{\Name{Tobias Schrank} \Email{tobias.schrank@tugraz.at}\\
  \Name{Franz Pernkopf} \Email{pernkopf@tugraz.at}\\
  \addr Graz University of Technology}

\begin{document}
\maketitle

\begin{abstract}
  Abstraction is a fundamental part when learning behavioral models
  of systems.  Usually the process of abstraction is manually defined
  by domain experts.  This paper presents a method to perform
  automatic abstraction for network protocols.  In particular a weakly
  supervised clustering algorithm is used to build an abstraction with
  a small vocabulary size for the widely used TLS protocol.  To show
  the effectiveness of the proposed method we compare the resultant
  abstract messages to a manually constructed (reference) abstraction.
  With a small amount of side-information in the form of a few
  labeled examples this method finds an abstraction that matches the
  reference abstraction perfectly.
\end{abstract}

\section{Introduction}
\label{sec:intro}

The implementation of complex designs is challenging even for experts.
This complexity often leads to bugs in the implementation.  Therefore,
complex pieces of software or hardware are usually extensive tested
and/or verified.  This, however, is complicated by the fact that most
designs are not formally specified, if at all.  Due to this
model-based testing with \emph{learned} models has become rather
popular in the recent past (e.g., \citet{deRuiterPoll:2015}).

However, the methods used in the field of learning-based testing --
namely a family of algorithms all building upon L*
\citep{Angluin:1987} -- rely on 
small alphabets in order to work in practice.  For instance,
\citet{SmeenkETAL:2015} make use of detailed knowledge of the system
under test to limit the size of the alphabet in order to make it
feasible to apply L*.
This reduction of the alphabet size is brought about by defining some
abstraction which is typically done \emph{by hand}.  This process,
however, relies on a considerable degree of knowledge of the system under
test that is not necessarily available.

In this paper we set forth to perform abstraction of the Transport
Layer Security (TLS) protocol with as little human intervention as
possible.  To this end, we employ a weakly supervised learning
algorithm from the $k$-means family that takes only a few labeled
examples as input.  It uses this kind of side-information to build a
set of constraints of which data points need to be assigned to the
same cluster and which must not.  These constraints are then in turn
used to both guide the algorithm towards both an optimal cluster
assignment and learning a distance metric with which the clusters can
be optimally separated.
We evaluate this semi-automatic method by comparing the cluster
assignments to a reference abstraction.  This reference abstraction is
of the form as a human would have to provide one when using common
methods. With a small amount of side-information in the form of a few
labeled examples this method finds an abstraction identical to the 
reference abstraction.


The choice to perform automatic abstraction of TLS is the direct
consequence of critical position TLS takes up in modern internet-based
communication. On top of this semi-automatic abstraction one can learn automata
that can be used in model-based testing -- a practice known as
learning-based testing.

This paper is structured as follows: In Section~\ref{sec:method} we present the semi-supervised clustering
algorithm. In Section~\ref{sec:exp} we discuss the experimental setup, the data
used and the evaluation procedures employed. Section~\ref{sec:results} reports results obtained from the
experiments and discusses their implications for the task of (semi-)automatic abstraction.
We conclude with a general discussion in Section~\ref{sec:conclusion}.

\setcounter{equation}{1}
\section{Method: MPCK-means}
\label{sec:method}

Metric learning and pairwise-constrained $k$-means (MPCK-means)
clustering \citep{BilenkoBasuMooney:2004} is an approach to
semi-supervised learning which employs pairwise constraints.  Pairwise
constraints come in two forms: \emph{must-link} constraints and
\emph{cannot-link} constraints.  MPCK-means employs these pairwise
constraints for both learning a metric space which separates data
points of cannot-link pairs and brings data points of must-link pairs
closer together as well as avoids constraint violations when assigning
data points to clusters.

MPCK-means employs the following objective function in
(\ref{eq:mpck-objective}).  It minimizes cluster dispersion under the
learned metric and at the same time reduces constraint violations:

\begin{align*}
  J_{mpckm} &= \sum_{\mathbf{x}_i \in \mathcal{X}}
              (d(\mathbf{x}_i, \boldsymbol{\mu}_{l_i})^2_{A_{l_i}})
    - \log(\det(A_{l_i})) \\
    &+ \sum_{(\mathbf{x}_i, \mathbf{x}_j) \in \mathcal{M}} w
    f_{\mathcal{M}}(\mathbf{x}_i, \mathbf{x}_j)
    \mathds{1}[l_i \neq l_j ] \\
    &+ \sum_{(\mathbf{x}_i, \mathbf{x}_j) \in \mathcal{C}} \bar{w}
    f_{\mathcal{C}} (\mathbf{x}_i, \mathbf{x}_j)
    \mathds{1} [l_i = l_j],
  \tag{\theequation}
  \label{eq:mpck-objective}
\end{align*}
where $\mathcal{X}$ is the set of data points, $l$ is the cluster
assignment of the current data point, $d(\cdot, \cdot )_{A_{l_i}}$ is the
current metric for this particular cluster, that is a weight matrix,
$w$ and $\bar{w}$ are the penalties imposed on violations of must-link
constraints and cannot-link constraints, respectively, $\mathcal{M}$
and $\mathcal{C}$ are the sets of must-link constraints and
cannot-link constraints, respectively, $f_{\mathcal{M}}$ and
$f_{\mathcal{C}}$ are functions defining the penalty imposed
violations depending on the closeness of the involved data points
under the current metric and $\mathds{1}$ is the indicator function.

The first term of objective (Eq.\,\ref{eq:mpck-objective}) accounts for
the dispersion given the current metrics $A_l$ (one for each
cluster).  The second term is needed as normalization of the magnitude
of the weights in $A_l$.  The third and forth term measures the number
of violated constraints weighted by a hyper-parameter ($w_{ij}$ and
$\bar{w}_{ij}$, respectively) and the distance between the two
involved data points.  Therefore, violated must-link constraints which
are considered to be far away under the current metrics $A_{l_i}$ and
$A_{l_j}$ result in larger penalties:

\begin{align}
  f_{\mathcal{M}} (\mathbf{x}_i, \mathbf{x}_j) = \frac{1}{2} d(\mathbf{x}_i, \mathbf{x}_j)^2_{A_{l_i}}
                             + \frac{1}{2} d(\mathbf{x}_i, \mathbf{x}_j)^2_{A_{l_j}}
\end{align}

Similarly, violated cannot-link constraints which are considered to be
close together under the current metric $A_{l_i}$ result in larger
penalties:

\begin{align}
  f_{\mathcal{C}} (\mathbf{x}_i, \mathbf{x}_j) = d(\mathbf{x}'_{l_i}, \mathbf{x}''_{l_j})^2 - d(\mathbf{x}_i, \mathbf{x}_j)^2_{A_{l_i}}
\end{align}
where $(\mathbf{x}'_{l_i}, \mathbf{x}''_{l_i})$ are the two maximally separated data
points according to the current metric $A_{l_i}$. This ensures that $f_{\mathcal{C}} (\mathbf{x}_i, \mathbf{x}_j)$ is non-negative.

For the experiments we use -- in contrast to the MPCK-mean's original
formulation -- a weighted hamming distance for $d(\cdot, \cdot )_{A_{l_i}}$.  Therefore, the weighting
matrix $A_{l_i}$  is diagonal.  In this sense employing
a diagonal weighting matrix corresponds to feature weighting.  



\section{Experiments}
\label{sec:exp}

All experiments presented in this text are implemented with the R
programming language \citep{R} and use TLS traces gathered with the
logging component of nqsb-tls \citep{nqsb-tls:2015}, a modern TLS
implementation.  In total, the data consists of approximately
53k~decoded TLS traces comprising nearly 370k~messages.  There are
just over 81k~unique messages. In this work, we used a random sub-set of 5k~messages where each message is truncated to a maximum of 32 fields (uninformative fields RANDOM and SESSIONID are filtered).
A sample TLS message is given in Table~\ref{fig:tls-sample}.

\begin{table}[htb]
  \centering
  \begin{tabular}{l l l}
    HANDSHAKE-IN & CLIENTHELLO \\
    VERSION & TLS\_1\_2 \\
    CIPHERSUITES & TLS\_RSA\_WITH\_AES\_128\_CBC\_SHA256 & \dots \\
    \vdots \\
  \end{tabular}
  \caption{A sample TLS message (decoded and truncated).}
  \label{fig:tls-sample}
\end{table}

\subsection{The TLS protocol}
\label{sec:tls}

One of today's most widely used cryptographic security protocol is
Transport Layer Security (TLS), predecessors of which are also known
as Secure Sockets Layer (SSL).  It is used to secure communications over
insecure channels in applications as diverse web browsing (as HTTPS),
email (as SMTPS), voice-over-IP (in SIP) and virtual private networks
(in OpenVPN).  For a long time, TLS implementations had been considered very secure
\cite{openvpn:whytls}.
A number of high-profile vulnerabilities starting in 2014 have changed
this picture, most famously a security bug called \emph{Heartbleed}.
%
Most TLS implementations provide a wide array of TLS versions,
protocol extensions, authentication modes and key exchange modes in
order to be maximally compatible with clients.

For this text's endeavor the interesting part of TLS is the 
process of establishing a secure connection between two nodes.  In
this process three sub-protocols are in use: To establish session
parameters and cryptographic keys the Handshake protocol is used.
Authentication, if asked for, is also done by means of this
sub-protocol.  To start the session with the keys established via the
handshake sub-protocol the ChangeCipherSpec sub-protocol is used.  To
signal errors or warnings to the other node the Alert sub-protocol is
used.

\subsection{Evaluation Measure}
\label{sec:eval}
In order to evaluate the quality of the automatic abstraction we
manually construct a \emph{reference abstraction} which is generated
by a set of hand-written rules.  This reference abstraction consists of $J=21$ classes and is similar
to the abstraction found in the literature (cf.\ for instance
\citet{BeurdoucheETAL:2015} or \citet{deRuiterPoll:2015}).

To measure the mismatch between the automatic abstraction and the
reference abstraction we employ two common measures for clustering
performance, namely, \emph{purity} and the \emph{adjusted Rand index} (ARI).  

For determining \emph{Purity} each cluster of the automatic abstraction is assigned to the reference abstraction where the intersection of samples is maximal. The number of intersecting samples is accumulated over all clusters and normalized \citep{ManningRaghavanSchuetze:2008}. It is formally defined as
$$
purity(\Omega, \mathbb{C}) = \frac{1}{N} \sum_k \max_j | \omega_k
\cap c_j |,
$$
where $N$ is number of samples, $\Omega=\left\{\omega_1, \omega_2, ...,\omega_K\right\}$ is the set of clusters provided by the algorithm, $\mathbb{C}=\left\{c_1, c_2, ..., C_J\right\}$ is the set of classes of the reference abstraction, $\omega_k$ denotes the set of samples in cluster $k$ and $c_j$ is the set of samples of the reference abstraction in class $j$. Purity is thus closely
related to \emph{accuracy} and a good indicator of goodness
for practical applications.

We also measure quality of abstraction through the ARI \citep{HubertArabie:1985}, a measure for similarity between two
partitioning that is adjusted for chance.  
%
The ARI equals $0$ if the clustering equals the expected value, and
$1$ for a perfect clustering.  Moreover, the ARI is geared more
towards a fair evaluation of methods than purity.

\subsection{Results and Discussion}
\label{sec:results}

In the present experiments we envision a scenario where constructing a
manual abstraction is considerably more costly than having a domain
expert deciding on whether two TLS messages should be considered equal
or unequal for the task at hand.

\begin{figure}[htb]
  \centering
  \subfigure[]{
    \includegraphics[width=.45\textwidth]{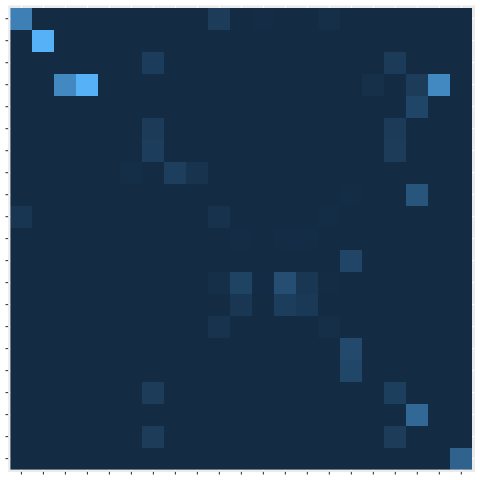}
  }
  \subfigure[]{
    \includegraphics[width=.45\textwidth]{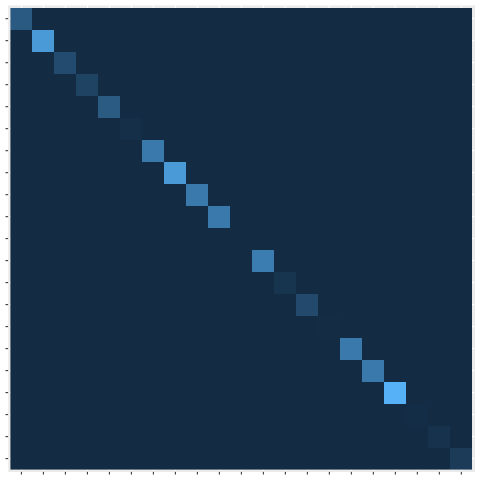}
  }
\caption{Confusion matrix: (a) Confusion matrix for baseline (unsupervised
  \emph{k}-means). 69\% purity, 48\% ARI. (b) Confusion matrix for MPCK-means; 5 labeled samples per class are used for the must-link and the cannot-link constraints; 100\% purity, 100\% ARI.
  Dark spots in the main diagonal correspond to small cluster sizes, not errors. }
\label{figA}
\end{figure}

The confusion matrix of the classical \emph{k}-means algorithm (unsupervised) is shown in Figure~\ref{figA}(a). It achieves a purity of 0.69 and an ARI of 0.48. In this experiment $K$ is set equal to $J=21$. Figure~\ref{figA}(b) shows the results for the MPCK-means using $K=J$. Additionally MPCK-means uses 5 labeled samples per class, i.e. in total 105 labels samples are used. Note that from these samples the must-link and cannot-link constraints can be formed. In this setting, we obtain an abstraction that matches the reference abstraction perfectly.

\begin{figure}[htb]
  \centering
  \subfigure[]{
    \includegraphics[width=.45\textwidth]{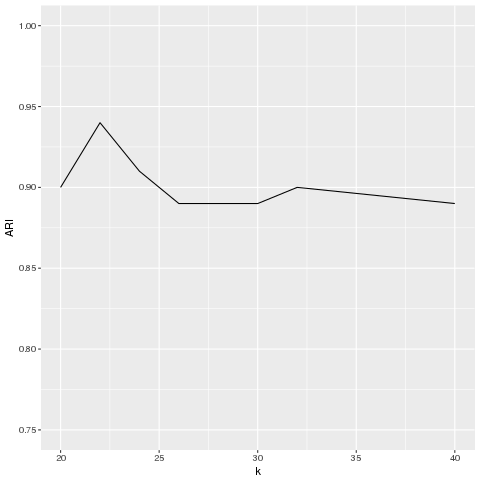}
  }
  \subfigure[]{
    \includegraphics[width=.45\textwidth]{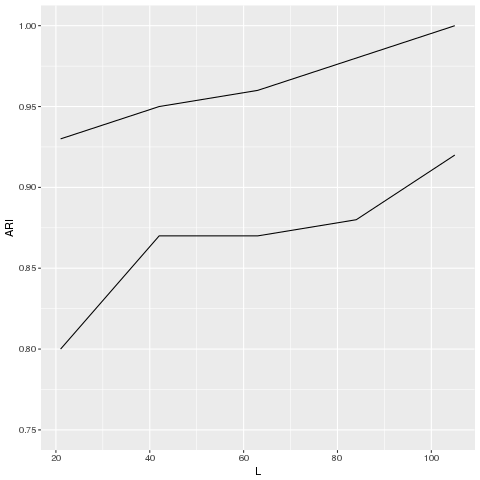}
  }
\caption{MPCK-means: (a) Variation of number of clusters $K$; 1 labeled sample per class is used for the cannot-link constraints; no must-link constraints can be formed. (b) Variation of number of labels per class (top: balanced, bottom:  unbalanced). }
\label{figB}
\end{figure}

In Figure~\ref{figB}(a) we show results for MPCK-means by varying $K\in\left\{20, \ldots, 40\right\}$. The best ARI is obtained for $K=22$. In this experiment only one labeled sample per class is used. Hence, only cannot-link constraints can be formed.
In Figure~\ref{figB}(b) we show results for MPCK-means by varying the number of labels per class from 1 to 5. The best ARI is obtained for 5 labels. In this experiment $K$ equals $J$. We achieve an ARI of 1 when using 5 labeled samples per class. In this case the cluster assignment of MPCK means matches the reference abstraction.

\section{Related work}
\label{sec:relwork}

\citet{WhalenBishopCrutchfield:2010} learn hidden Markov models of the
message format of text-based network protocols (HTTP, FTP).%
\citet{CuiKannanWang:2007} apply recursive clustering and merging to
learn the format of both text-based and binary network protocols
(HTTP, RPC, SMB).

Some research in this field assume additionally access to the software
binary in order to perform dynamic binary analysis, i.e. analyze the
interaction between the executing system and the executed software
\citep{CaballeroETAL:2007,WondracekETAL:2008,WangETAL:2009,ComparettiETAL:2009}.
This condition is not met in the scenario envisioned in this paper
where we only assume access to the system under test over the network.

In a different line of research, \citep{AartsETAL:2012} integrate the
task of finding valid abstractions into the loop of learning a finite
state machine of the system under test.

\section{Conclusion}
\label{sec:conclusion}

In this paper, we investigated the semi-automatic abstraction of the
network protocol TLS. In particular, we use a small amount of side-information in the form of a few
labeled examples. This enables to find an abstraction that matches the reference abstraction perfectly.

In future work, we aim to use our automatic abstraction to feed an automaton learning algorithm to show its usefulness in frameworks currently in use in model-based testing. Furthermore, we plan to use MPCK-means for other protocols beyond TLS.

\acks{This work was supported by the LEAD Project “Dependable Internet of Things” funded by Graz University of Technology.}

\bibliography{learnaut}

\end{document}